# Learning Rich Features for Image Manipulation Detection

Peng Zhou[1]  Xintong Han[1]  Vlad I. Morariu[2] *  Larry S. Davis[1]
[1]University of Maryland, College Park  [2]Adobe Research
pengzhou@umd.edu {xintong,lsd}@umiacs.umd.edu morariu@adobe.com

## Abstract

*Image manipulation detection is different from traditional semantic object detection because it pays more attention to tampering artifacts than to image content, which suggests that richer features need to be learned. We propose a two-stream Faster R-CNN network and train it end-to-end to detect the tampered regions given a manipulated image. One of the two streams is an RGB stream whose purpose is to extract features from the RGB image input to find tampering artifacts like strong contrast difference, unnatural tampered boundaries, and so on. The other is a noise stream that leverages the noise features extracted from a steganalysis rich model filter layer to discover the noise inconsistency between authentic and tampered regions. We then fuse features from the two streams through a bilinear pooling layer to further incorporate spatial co-occurrence of these two modalities. Experiments on four standard image manipulation datasets demonstrate that our two-stream framework outperforms each individual stream, and also achieves state-of-the-art performance compared to alternative methods with robustness to resizing and compression.*

## 1. Introduction

With the advances of image editing techniques and user-friendly editing software, low-cost tampered or manipulated image generation processes have become widely available. Among tampering techniques, splicing, copy-move, and removal are the most common manipulations. Image splicing copies regions from an authentic image and pastes them to other images, copy-move copies and pastes regions within the same image, and removal eliminates regions from an authentic image followed by inpainting. Sometimes, post-processing like Gaussian smoothing will be applied after these tampering techniques. Examples of these manipulations are shown in Figure 1. Even with careful inspection, humans find it difficult to recognize the tampered regions.

---

*The work was done while the author was at the University of Maryland

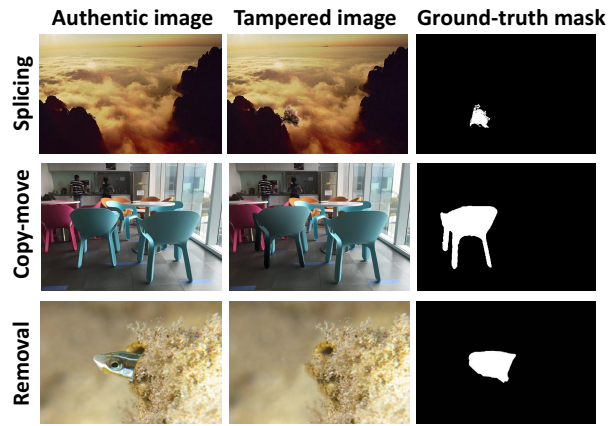

Figure 1. Examples of tampered images that have undergone different tampering techniques. From the top to bottom are the examples showing manipulations of splicing, copy-move and removal.

As a result, distinguishing authentic images from tampered images has become increasingly challenging. The emerging research focusing on this topic — image forensics — is of great importance because it seeks to prevent attackers from using their tampered images for unscrupulous business or political purposes. In contrast to current object detection networks [28, 18, 10, 32, 16, 31] which aim to detect all objects of different categories in an image, a network for image manipulation detection would aim to detect only the tampered regions (usually objects). We investigate how to adopt object detection networks to perform image manipulation detection by exploring both RGB image content and image noise features.

Recent work on image forensics utilizes clues such as local noise features [35, 26] and Camera Filter Array (CFA) patterns [19] to classify a specific patch or pixel [11] in an image as tampered or not, and localize the tampered regions [19, 9, 6]. Most of these methods focus on a single tampering technique. A recently proposed architecture [2] based on a Long Short Term Network (LSTM) segments tampered patches, showing robustness to multiple tampering tech-niques by learning to detect tampered edges. Here, we propose a novel two-stream manipulation detection framework, which not only models visual tampering artifacts (*e.g.*, tampered artifacts near manipulated edges), but also captures inconsistencies in local noise features.

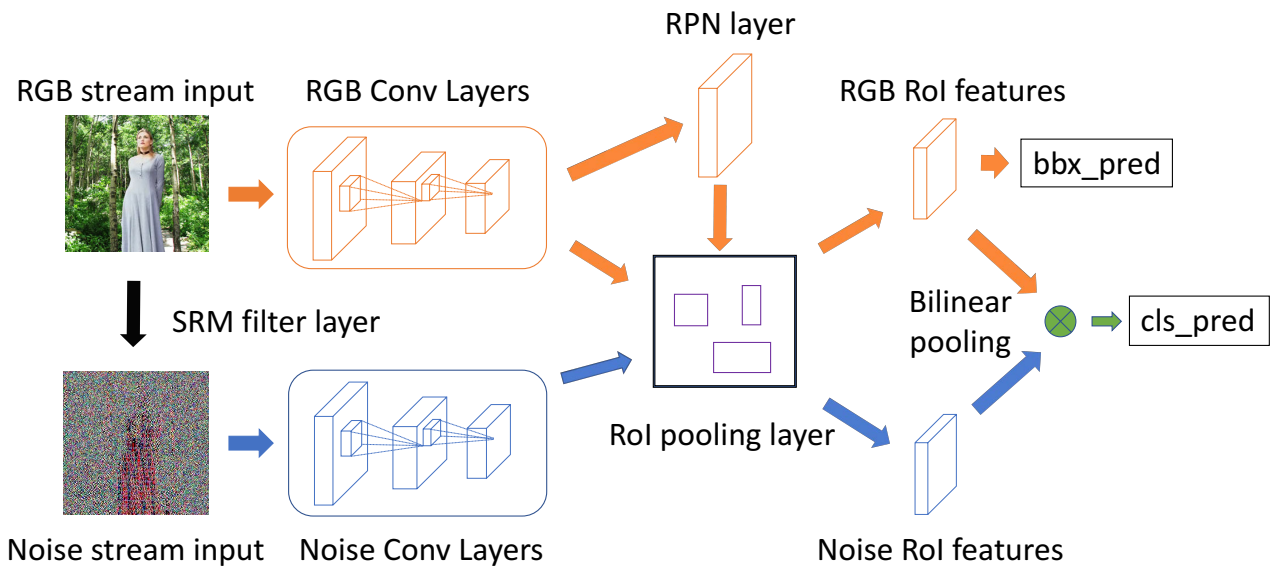

Figure 2. Illustration of our two-stream Faster R-CNN network. The RGB stream models visual tampering artifacts, such as unusually high contrast along object edges, and regresses bounding boxes to the ground-truth. The noise stream first obtains the noise feature map by passing input RGB image through an SRM filter layer, and leverages the noise features to provide additional evidence for manipulation classification. The RGB and noise streams share the same region proposals from RPN network which only uses RGB features as input. The RoI pooling layer selects spatial features from both RGB and noise streams. The predicted bounding boxes (denoted as 'bbx_pred') are generated from RGB RoI features. A bilinear pooling [23, 17] layer after RoI pooling enables the network to combine the spatial co-occurrence features from the two streams. Finally, passing the results through a fully connected layer and a softmax layer, the network produces the predicted label (denoted as 'cls_pred') and determines whether predicted regions have been manipulated or not.

More specifically, we adopt Faster R-CNN [28] within a two-stream network and perform end-to-end training. A summary of our method is shown in Figure 2. Deep learning detection models like Faster R-CNN [28] have demonstrated good performance on detecting semantic objects over a range of scales. The Region Proposal Network (RPN) is the component in Faster R-CNN that is responsible for proposing image regions that are likely to contain objects of interest, and it can be adapted for image manipulation detection. For distinguishing tampered regions from authentic regions, we utilize features from the RGB channels to capture clues like visual inconsistencies at tampered boundaries and contrast effect between tampered regions and authentic regions. The second stream analyzes the local noise features in an image.

The intuition behind the second stream is that when an object is removed from one image (the source) and pasted into another (the target), the noise features between the source and target images are unlikely to match. These differences can be partially masked if the user subsequently compresses the tampered image [26, 4]. To utilize these features, we transform the RGB image into the noise domain and use the local noise features as the input to the second stream. There are many ways to produce noise features from an image. Based on recent work on steganalysis rich model (SRM) for manipulation classification [35, 15], we select SRM filter kernels to produce the noise features and use them as the input channel to the second Faster R-CNN network.

Features from these two streams are then bi-linearly pooled for each Region of Interest (RoI) to detect tampering artifacts based on features from both streams, see Figure 2.

Previous image manipulation datasets [25, 1, 12, 30] contain only several hundred images, not enough to train a deep network. To overcome this, we created a synthetic tampering dataset based on COCO [22] for pre-training our model and then finetuned the model on different datasets for testing. Experimental results of our approach on four standard datasets demonstrate promising performance.

Our contribution is two-fold. First, we show how a Faster R-CNN framework can be adapted for image manipulation detection in a two-stream fashion. We explore two modalities, RGB tampering artifacts and local noise feature in-consistencies, bilinearly pooling them to identify tampered regions. Second, we show that the two streams are complementary for detecting different tampered techniques, leading to improved performance on four image manipulation datasets compared to state-of-the-art methods.

## 2. Related Work

Research on image forensics consists of various approaches to detect the low-level tampering artifacts within a tampered image, including double JPEG compression [4], CFA color array anayIsis [19] and local noise analysis [7]. Specifically, Bianchi *et al*. [4] propose a probabilistic model to estimate the DCT coefficients and quantization factors for different regions. CFA based methods analyze low-level statistics introduced by the camera internal filter patterns under the assumption that the tampered regions disturb these patterns. Goljan *et al*. [19] propose a Gaussian Mixture Model (GMM) to classify CFA present regions (authentic regions) and CFA absent regions (tampered regions).

Recently, local noise features based methods, like the steganalysis rich model (SRM) [15], have shown promising performance in image forensics tasks. These methods extract local noise features from adjacent pixels, capturing the inconsistency between tampered regions and authentic regions. Cozzolino *et al*. [7] explore and demonstrate the performance of SRM features in distinguishing tampered and authentic regions. They also combine SRM features by including the quantization and truncation operations with a Convolutional Neural Network (CNN) to perform manipulation localization [8]. Rao *et al*. [27] use an SRM filter kernel as initialization for a CNN to boost the detection accuracy. Most of these methods focus on specific tampering artifacts and are limited to specific tampering techniques. We also use these SRM filter kernels to extract low-level noise that is used as the input to a Faster R-CNN network, and learn to capture tampering traces from the noise features. Moreover, a parallel RGB stream is trained jointly to model mid- and high-level visual tampering artifacts.

With the success of deep learning techniques in various computer vision and image processing tasks, a number of recent techniques have also employed deep learning to address image manipulation detection. Chen *et al*. [5] add a low pass filter layer before a CNN to detect median filtering tampering techniques. Bayar *et al*. [3] change the low pass filter layer to an adaptive kernel layer to learn the filtering kernel used in tampered regions. Beyond filtering learning, Zhang *et al*. [34] propose a stacked autoencoder to learn context features for image manipulation detection. Cozzolino *et al*. [9] treat this problem as an anomaly detection task and use an autoencoder based on extracted features to distinguish those regions that are difficult to reconstruct as tampered regions. Salloum *et al*. [29] use a Fully Convolutional Network (FCN) framework to directly predict the tampering mask given an image. They also learn a boundary mask to guide the FCN to look at tampered edges, which assists them in achieving better performance in various image manipulation datasets. Bappy *et al*. [2] propose an LSTM based network applied to small image patches to find the tampering artifacts on the boundaries between tampered patches and image patches. They jointly train this network with pixel level segmentation to improve the performance and show results under different tampering techniques. However, only focusing on nearby boundaries provides limited success in different scenarios, *e.g.*, removing the whole object might leave no boundary evidence for detection. Instead, we use global visual tampering artifacts as well as the local noise features to model richer tampering artifacts. We use a two-stream network built on Faster R-CNN to learn rich features for image manipulation detection. The network shows robustness to splicing, copy-move and removal. In addition, the network enables us to make a classification of the suspected tampering techniques.

## 3. Proposed Method

We employ a multi-task framework that simultaneously performs manipulation classification and bounding box regression. RGB images are provided in the RGB stream (the top stream in Figure 2), and SRM images in the noise stream (the bottom stream in Figure 2). We fuse the two streams through bilinear pooling before a fully connected layer for manipulation classification. The RPN uses the RGB stream to localize tampered regions.

### 3.1. RGB Stream

The RGB stream is a single Faster R-CNN network and is used both for bounding box regression and manipulation classification. We use a ResNet 101 network [20] to learn features from the input RGB image. The output features of the last convolutional layer of ResNet are used for manipulation classification.

The RPN network in the RGB stream utilizes these features to propose RoI for bounding box regression. Formally, the loss for the RPN network is defined as

$$L_{RPN}(g_i, f_i) = \frac{1}{N_{cls}} \sum_i L_{cls}(g_i, g_i^\star)$$
$$+ \lambda \frac{1}{N_{reg}} \sum_i g_i^\star L_{reg}(f_i, f_i^\star), \quad (1)$$

where $g_i$ denotes the probability of anchor $i$ being a potential manipulated region in a mini batch, and $g_i^\star$ denotes the ground-truth label for anchor $i$ to be positive. The terms $f_i$, $f_i^\star$ are the 4 dimensional bounding box coordinates for anchor i and the ground-truth, respectively. $L_{cls}$ denotes cross entropy loss for RPN network and $L_{reg}$ denotes smooth $L_1$ loss for regression for the proposal bounding boxes. $N_{cls}$ denotes the size of a mini-batch in the RPN network. $N_{reg}$ is the number of anchor locations. The term $\lambda$ is a hyper-parameter to balance the two losses and is set to 10. Note that in contrast to traditional object detection whose RPN network searches for regions that are likely to be objects, our RPN network searches for regions that are likely to be manipulated. The proposed regions might not necessarily be objects, *e.g.*, the case in the removal tampering process.

### 3.2. Noise Stream

RGB channels are not sufficient to tackle all the different cases of manipulation. In particular, tampered images that were carefully post processed to conceal the splicing boundary and reduce contrast differences are challenging for the RGB stream.

So, we utilize the local noise distributions of the image to provide additional evidence. In contrast to the RGB stream, the noise stream is designed to pay more attention to noise rather than semantic image content. This is novel — while current deep learning models do well in representing hierarchical features from RGB image content, no prior work in deep learning has investigated learning from noise distributions in detection. Inspired by recent progress on SRM features from image forensics [15], we use SRM filters to extract the local noise features (examples shown in Figure 3) from RGB images as the input to our noise stream.

In our setting, noise is modeled by the residual between a pixel's value and the estimate of that pixel's value produced by interpolating only the values of neighboring pixels. Starting from 30 basic filters, along with nonlinear operations like maximum and minimum of the nearby outputs after filtering, SRM features gather the basic noise features. SRM quantifies and truncates the output of these filters and extracts the nearby co-occurrence information as the final features. The feature obtained from this process can be regarded as a local noise descriptor [7]. We find that only using 3 kernels can achieve decent performance, and applying all 30 kernels does not give significant performance gain. Therefore, we choose 3 kernels, whose weights are shown in Figure 4, and directly feed these into a pre-trained network trained on 3-channel inputs. We define the kernel size of the SRM filter layer in the noise stream to be $5 \times 5 \times 3$. The output channel size of our SRM layer is 3.

The resulting noise feature maps after the SRM layer are shown in the third column of Figure 3. It is clear that they emphasize the local noise instead of image content and explicitly reveal tampering artifacts that might not be visible in the RGB channels. We directly use the noise features as the input to the noise stream network. The backbone convolutional network architecture of the noise stream is the same as the RGB stream. The noise stream shares the same RoI pooling layer as the RGB stream. For bounding box regression, we only use the RGB channels because RGB features perform better than noise features for the RPN network based on our experiments (See Table 1).

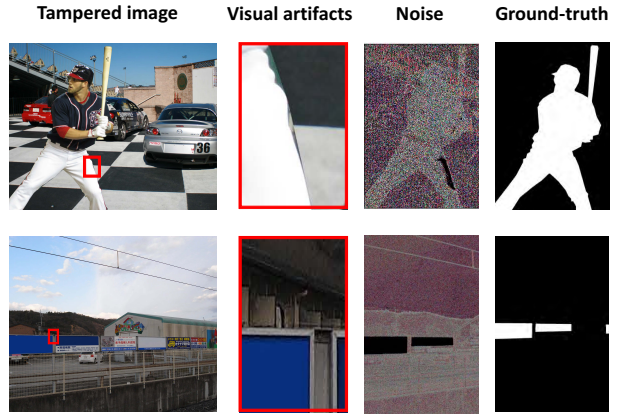

Figure 3. Illustration of tampering artifacts. Two examples showing tampering artifacts in the original RGB image and in the local noise features obtained by the SRM filter layer. The second column is the amplified regions for the red bounding boxes in the first column. As shown in the second column, the unnaturally high contrast along the baseball player's edges provides a strong cue about the presence of tampering. The third column shows the local noise inconsistency between tampered regions and authentic regions. In different scenarios, visual information and noise features play a complementary role in revealing tampering artifacts.

$$\frac{1}{4}\begin{bmatrix} 0 & 0 & 0 & 0 & 0 \\ 0 & -1 & 2 & -1 & 0 \\ 0 & 2 & -4 & 2 & 0 \\ 0 & -1 & 2 & -1 & 0 \\ 0 & 0 & 0 & 0 & 0 \end{bmatrix} \quad \frac{1}{12}\begin{bmatrix} -1 & 2 & -2 & 2 & -1 \\ 2 & -6 & 8 & -6 & 2 \\ -2 & 8 & -12 & 8 & -2 \\ 2 & -6 & 8 & -6 & 2 \\ -1 & 2 & -2 & 2 & -1 \end{bmatrix} \quad \frac{1}{2}\begin{bmatrix} 0 & 0 & 0 & 0 & 0 \\ 0 & 0 & 0 & 0 & 0 \\ 0 & 1 & -2 & 1 & 0 \\ 0 & 0 & 0 & 0 & 0 \\ 0 & 0 & 0 & 0 & 0 \end{bmatrix}$$

Figure 4. The three SRM filter kernels used to extract noise features.

### 3.3. Bilinear Pooling

We finally combine the RGB stream with the noise stream for manipulation detection. Among various fusion methods, we apply bilinear pooling on features from both streams. Bilinear pooling [23], first proposed for fine-grained classification, combines streams in a two-stream CNN network while preserving spatial information to improve the detection confidence. The output of our bilinear pooling layer is $x = f_{RGB}^T f_N$, where $f_{RGB}$ is the RoI feature of the RGB stream and $f_N$ is the RoI feature of the noise stream. Sum pooling squeezes the spatial feature before classification. We then apply signed square root ($x \leftarrow sign(x)\sqrt{|x|}$) and $L_2$ normalization before forwarding to the fully connected layer.

To save memory and speed up training without decreasing performance, we use compact bilinear pooling as proposed in [17].

After the fully connected and softmax layers, we obtain the predicted class of the RoI regions, as indicated in Figure 2. We use cross entropy loss for manipulation classification and smooth $L_1$ loss for bounding box regression. The total loss function is:

$$L_{total} = L_{RPN} + L_{tamper}(f_{RGB}, f_N) + L_{bbox}(f_{RGB}), \quad (2)$$

where $L_{total}$ denotes total loss. $L_{RPN}$ denotes the RPN loss in RPN network. $L_{tamper}$ denotes the final cross entropy classification loss, which is based on the bilinear pooling feature from both the RGB and noise stream. $L_{bbox}$ denotes the final bounding box regression loss. $f_{RGB}$ and $f_N$ are the RoI features from RGB and noise streams. The summation of all terms produces the total loss function.

### 3.4. Implementation Detail

The proposed network is trained end-to-end. The input image as well as the extracted noise features are re-sized so that the shorter length equals to 600 pixels. Four anchor scales with size from $8^2$, $16^2$, $32^2$ to $64^2$ are used, and the aspect ratios are 1:2, 1:1 and 2:1. The feature size after RoI pooling is $7 \times 7 \times 1024$ for both RGB and noise streams. The output feature size of compact bilinear pooling is set to 16384. The batch size of RPN proposal is 64 for training and 300 for testing.

Image flipping is used for data augmentation. The Intersection-over Union (IoU) threshold for RPN positive example (potential manipulated regions) is 0.7 and 0.3 for negative example (authentic regions). Learning rate is initially set to 0.001 and then is reduced to 0.0001 after 40K steps. We train our model for 110K steps. At test time, standard Non-Maximum Suppression (NMS) is applied to reduce the redundancy of proposed overlapping regions. The NMS threshold is set to 0.2.

## 4. Experiments

We demonstrate our two stream network on four standard image manipulation datasets and compare the results with state-of-the-art methods. We also compare different data augmentations and measure the robustness of our method to resizing and JPEG compression.

### 4.1. Pre-trained Model

Current standard datasets do not have enough data for deep neural network training. To test our network on these datasets, we pre-train our model on our synthetic dataset. We automatically create a synthetic dataset using the images and annotations from COCO [22]. We use the segmentation annotations to randomly select objects from COCO [22], and then copy and paste them to other images. The training (90%) and testing set (10%) is split to ensure the same background and tampered object do not appear in both training and testing set. Finally, we create 42K tampered and authentic image pairs. We will release this dataset for research use. The output of our model is bounding boxes with confidence scores indicating whether the detected regions have been manipulated or not.

| AP | Synthetic test |
|---|---|
| RGB Net | 0.445 |
| Noise Net | 0.461 |
| RGB-N noise RPN | 0.472 |
| Noise + RGB RPN | 0.620 |
| RGB-N | 0.627 |

Table 1. AP comparison on our synthetic COCO dataset. The row is the model architectures, where RGB Net is a single Faster R-CNN using RGB image as input; Noise Net is a single Faster R-CNN using noise feature map as input; RGB-N noise RPN is a two-stream Faster R-CNN using noise features for RPN network. Noise + RGB RPN is a two-stream Faster R-CNN using both noise and RGB features as the input of RPN network. RGB-N is a two-stream Faster R-CNN using RGB features for RPN network.

To include some authentic regions in Region of Interest (RoI) for better comparison, We slightly enlarge the default bounding boxes by 20 pixels during training so that both the RGB and noise streams learn the inconsistency between tampered and authentic regions.

We train our model end-to-end on this synthetic dataset. The ResNet 101 used in Faster R-CNN is pre-trained on ImageNet. We use Average Precision (AP) for evaluation, the metric of which is the same as COCO [22] detection evaluation. We compare the result of the two-stream network with each one of the streams in Table 1. This table shows that our two-stream network performs better than each single stream. Also, the comparison among RGB-N, RGB-N using noise features as RPN and RPN uses both features shows that RGB features are more suitable than noise features to generate region proposals.

### 4.2. Testing on Standard Datasets

**Datasets**. We compare our method with current state-of-the-art methods on NIST Nimble 2016 [1] (NIST16), CASIA [12, 13], COVER [30] and Columbia dataset.
• NIST16 is a challenging dataset which contains all three tampering techniques. The manipulations in this dataset are post-processed to conceal visible traces. They also provide ground-truth tampering mask for evaluation.
• CASIA provides spliced and copy-moved images of various objects. The tampered regions are carefully selected and some post processing like filtering and blurring is also applied. Ground-truth masks are obtained by thresholding the difference between tampered and original images. We use CASIA 2.0 for training and CASIA 1.0 for testing.
• COVER is a relatively small dataset focusing on copy-move. It covers similar objects as the pasted regions to conceal the tampering artifacts (see the second row in Figure 1). Ground-truth masks are provided.
• Columbia dataset focuses on splicing based on uncompressed images. Ground-truth masks are provided.

| Datasets | NIST16 | CASIA | Columbia | COVER |
|---|---|---|---|---|
| Training | 404 | 5123 | - | 75 |
| Testing | 160 | 921 | 180 | 25 |

Table 2. Training and testing split (number of images) for four standard datasets. Columbia is only used for testing the model trained on our synthetic dataset.

To fine-tune our model on these datasets, we extract the bounding box from the ground-truth mask. We compare with other approaches on the same training and testing split protocol as [2] (for NIST16 and COVER) and [29] (for Columbia and CASIA). See Table 2.

**Evaluation Metric**. We use pixel level $F_1$ score and Area Under the receiver operating characteristic Curve (AUC) as our evaluation metrics for performance comparison. $F_1$ score is a pixel level evaluation metric for image manipulation detection, as discussed in [33, 29]. We vary different thresholds and use the highest $F_1$ score as the final score for each image, which follows the same protocol in [33, 29]. We assign the confidence score to every pixel in the detected bounding boxes for pixel-level AUC evaluation.

**Baseline Models**. We compare our proposed method with various baseline models as described below:
• ELA: An error level analysis method [21] which aims to find the compression error difference between tampered regions and authentic regions through different JPEG compression qualities.
• NOI1: A noise inconsistency based method using high pass wavelet coefficients to model local noise [24].
• CFA1: A CFA pattern estimation method [14] which uses nearby pixels to approximate the camera filter array patterns and then produces the tampering probability for each pixel.
• MFCN: A multi-task edge-enhanced FCN based network [29] jointly detecting tampered edges using edge binary masks and tampered regions using tampered region masks.
• J-LSTM: An LSTM based network [2] jointly training patch level tampered edge classification and pixel level tampered region segmentation.
• RGB Net: A single Faster R-CNN network with RGB images as input. *i.e.*, our RGB Faster R-CNN stream.
• Noise Net: A single Faster R-CNN network with noise feature map as input obtained from a SRM filter layer. The RPN network uses noise features in this case.
• Late Fusion: Direct fusion combining all detected bounding boxes for both RGB Net and noise Net. The confi-

|  | NIST16 | Columbia | COVER | CASIA |
|---|---|---|---|---|
| ELA [21] | 0.236 | 0.470 | 0.222 | 0.214 |
| NOI1 [24] | 0.285 | 0.574 | 0.269 | 0.263 |
| CFA1 [14] | 0.174 | 0.467 | 0.190 | 0.207 |
| MFCN [29] | 0.571 | 0.612 | - | **0.541** |
| RGB Net | 0.567 | 0.585 | 0.391 | 0.392 |
| Noise Net | 0.521 | **0.705** | 0.355 | 0.283 |
| Late Fusion | 0.625 | 0.681 | 0.371 | 0.397 |
| **RGB-N (ours)** | **0.722** | 0.697 | **0.437** | 0.408 |

Table 3. $F_1$ score comparison on four standard datasets. '-' denotes that the result is not available in the literature.

|  | NIST16 | Columbia | COVER | CASIA |
|---|---|---|---|---|
| ELA [21] | 0.429 | 0.581 | 0.583 | 0.613 |
| NOI1 [24] | 0.487 | 0.546 | 0.587 | 0.612 |
| CFA1 [14] | 0.501 | 0.720 | 0.485 | 0.522 |
| J-LSTM [2] | 0.764 | - | 0.614 | - |
| RGB Net | 0.857 | 0.796 | 0.789 | 0.768 |
| Noise Net | 0.881 | 0.851 | 0.753 | 0.693 |
| Late Fusion | 0.924 | 0.856 | 0.793 | 0.777 |
| **RGB-N (ours)** | **0.937** | **0.858** | **0.817** | **0.795** |

Table 4. Pixel level AUC comparison on four standard datasets. '-' denotes that the result is not available in the literature.

dence scores of the overlapping detected regions from the two streams are set to the maximum one.
• RGB-N: Bilinear pooling of RGB stream and noise stream for manipulation classification and RGB stream for bounding box regression. *i.e.* our full model.

We use the $F_1$ scores of NOI1, CFA1 and ELA reported in [29] and run the code provided by [33] to obtain the AUC results. The results of MFCN and J-LSTM are replicated from the original literatures as their code is not publicly available.

Table 3 shows the $F_1$ score comparison between our method and the baselines. Table 4 provides the AUC comparison. From these two tables, it is clear that our method outperforms conventional methods like ELA, NOI1 and CFA1. This is because they all focus on specific tampering artifacts that only contain partial information for localization, which limits their performance. Our approach outperforms MFCN on Columbia and NIST16 dataset.

One of the reasons our method achieves better performance than J-LSTM is that J-LSTM seeks tampered edges as evidence of tampering, which cannot always detect the entire tampered regions. Also, our method has larger receptive field and captures global context rather than nearby pixels, which helps collect more cues like contrast difference for manipulation classification.

As shown in Table 3 and 4, our RGB-N network also improves the individual streams for all the datasets except Columbia. Columbia only contains uncompressed spliced regions, which preserves noise differences so well that it is sufficient to use only the noise features. This yields satisfactory performance for the noise stream.

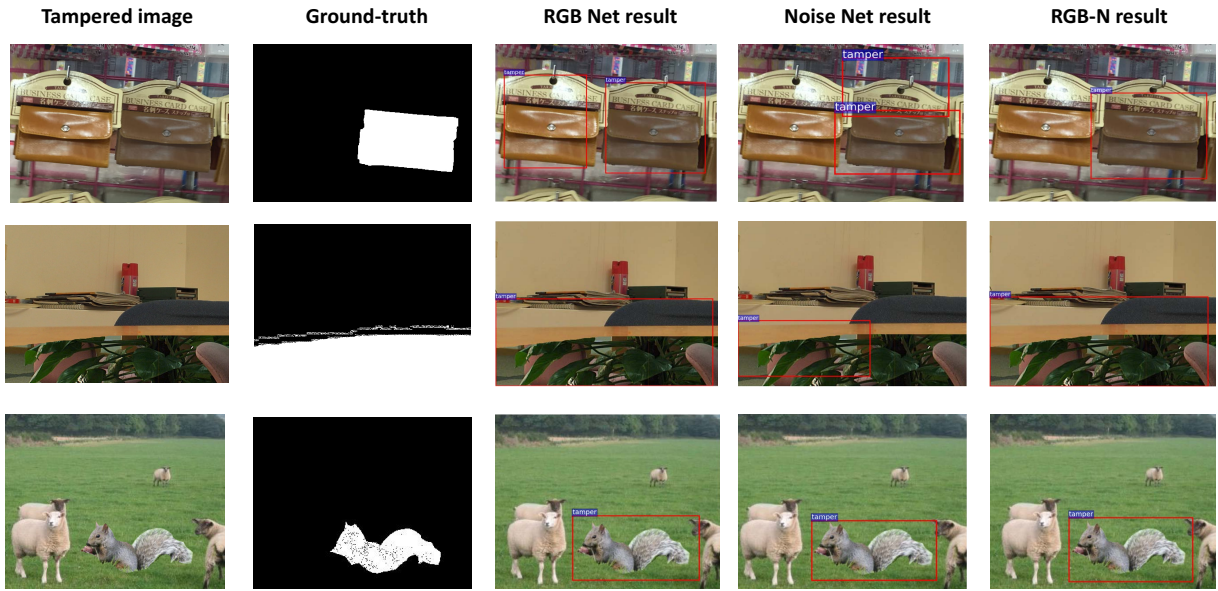

Figure 5. Qualitative visualization of results. The top row shows a qualitative result from the COVER dataset. The copy-moved bag confuses the RGB Net, and the noise Net. RGB-N achieves a better detection in this case because it combines the features from the two streams. The middle row shows a qualitative result from the Columbia. The RGB Net produces a more accurate result than noise stream. Taking into account both streams produces a better result for RGB-N. The bottom row shows a qualitative result from the CASIA1.0. The spliced object leaves clear tampering artifacts in both the RGB and noise streams, which yields precise detections for the RGB, noise, and RGB-N networks.

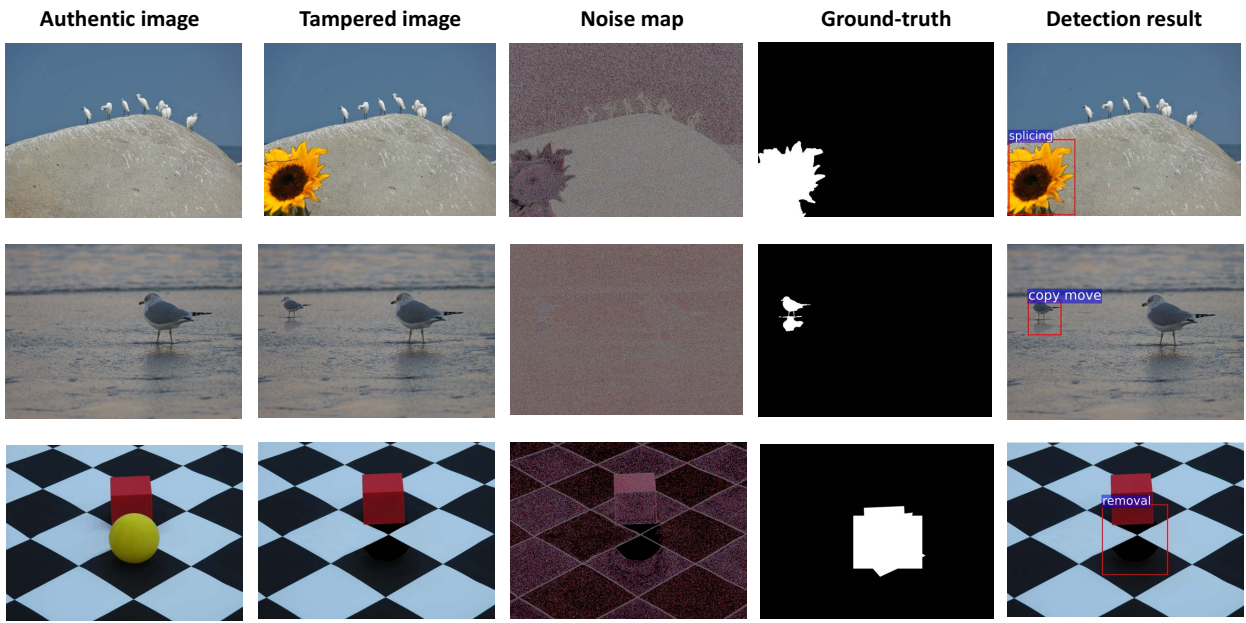

Figure 6. Qualitative results for multi-class image manipulation detection on NIST16 dataset. RGB and noise map provide different information for splicing, copy-move and removal. By combining the features from the RGB image with the noise features, RGB-N produces the correct classification for different tamepring techniques.

| $F_1$/AUC | NIST16 | COVER | CASIA |
|---|---|---|---|
| Flipping + JPEG | 0.712/**0.950** | 0.425/0.810 | **0.413**/0.785 |
| Flipping + noise | 0.717/0.947 | 0.412/0.801 | 0.396/0.776 |
| Flipping | **0.722**/0.937 | **0.437**/0.817 | 0.408/**0.795** |
| No flipping | 0.716/0.940 | 0.312/0.793 | 0.361/0.766 |

Table 5. Data augmentation comparison. Flipping: image flipping. JPEG: JPEG compression with quality 70. Noise: adding Gaussian noise with variance of 5. Each entry is $F_1$/AUC score.

| JPEG/Resizing | 100/1 | 70/0.7 | 50/0.5 |
|---|---|---|---|
| NOI1 | 0.285/0.285 | 0.142/0.147 | 0.140/0.155 |
| ELA | 0.236/0.236 | 0.119/0.141 | 0.114/0.114 |
| CFA1 | 0.174/0.174 | 0.152/0.134 | 0.139/0.141 |
| RGB-N | **0.722/0.722** | **0.677/0.689** | **0.677/0.681** |

Table 6. $F_1$ score on NIST16 dataset for JPEG compression (with quality 70 and 50) and resizing (with scale 0.7 and 0.5) attacks. Each entry is the $F_1$ score of JPEG/Resizing.

|  | Splicing | Removal | Copy-Move | Mean |
|---|---|---|---|---|
| AP | 0.960 | 0.939 | 0.903 | 0.934 |

Table 7. AP comparison on multi-class on NIST16 dataset using the RGB-N network. Mean denotes the mean AP for splicing, removal and copy-move.

For all datasets, late fusion performs worse than RGB-N, which shows the effectiveness of our fusion approach.

**Data Augmentation.** We compare different data augmentation methods in Table 5. Compared with no augmentation, image flipping improves the performance and other augmentation methods like JPEG compression and noise contribute little improvement.

**Robustness to JPEG and Resizing Attacks.** We test the robustness of our method and compare with 3 methods (whose code is available) in Table 6. Our method is more robust to these attacks and outperforms other methods.

### 4.3. Manipulation Technique Detection

The rich feature representation of our network enables it to distinguish between different manipulation techniques as well. We explore manipulation technique detection and analyze the detection performance for all three tampering techniques. NIST16 contains the labels for all three tampering techniques, which enables multi-class image manipulation detection. We change the classes for manipulation classification to be splicing, removal and copy-move so as to learn distinct visual tampering artifacts and noise features for each class. The performance of each tamper class is shown in Table 7.

The AP result in Table 7 indicates that splicing is the easiest manipulation techniques to detect using our method. This is because splicing has a high probability to produce both RGB artifacts like unnatural edges, contrast differences as well as noise artifacts. Removal detection performance also beats copy-move because the inpainting that follows the removal process has a large effect on the noise features, as shown in Figure 3. Copy-move is the most difficult tamper technique for our proposed method. The explanation is that on one hand, the copied regions are from the same image, which yields a similar noise distribution to confuse our noise stream. On the other hand, the two regions generally have the same contrast. Also, the technique would ideally need to compare the two objects to each other (*i.e.*, it would need to find and compare two RoIs at the same time), which the current approach does not do. Thus, our RGB stream has less evidence to distinguish between the two regions.

### 4.4. Qualitative Result

We show some qualitative results in Figure 5 for comparison of RGB, noise and RGB-N network in two-class image manipulation detection. The images are selected from the COVER, Columbia and CASIA 1.0. Figure 5 provides examples for which our two-stream network yields good performance even if one of the single streams fails (the first and second row in Figure 5).

Figure 6 shows the results of the RGB-N network on the task of manipulation technique detection task using the NIST16. As is shown in the figure, our network produces accurate results for different tampering techniques.

## 5. Conclusion

We propose a novel network using both an RGB stream and a noise stream to learn rich features for image manipulation detection. We extract noise features by an SRM filter layer adapted from steganalysis literatures, which enables our model to capture noise inconsistency between tampered and authentic regions. We explore the complementary contribution of finding tampered regions from RGB and the noise features of an image. Not surprisingly, the fusion of the two streams leads to improved performance. Experiments on standard datasets show that our method not only detects tampering artifacts but also distinguishes between various tampering techniques. More features, including JPEG compression, will be explored in the future.

## Acknowledgement

This work was supported by the DARPA MediFor program under cooperative agreement FA87501620191, "Physical and Semantic Integrity Measures for Media Forensics". The authors acknowledge the Maryland Advanced Research Computing Center (MARCC) for providing computing resources.